\documentclass[letterpaper, 10pt, journal, twoside]{IEEEtran}

\usepackage[utf8]{inputenc}
\usepackage{amssymb,amsfonts,amsmath,amscd}
\usepackage{bm}
\usepackage[pdftex]{graphicx}
\usepackage{url}
\usepackage{cite}
\usepackage[usenames,dvipsnames]{color}
\usepackage{subcaption}
\usepackage{caption}
\usepackage{accents}
\usepackage{tabularx}
\usepackage{tikz}
\usepackage{stackengine}
\usepackage{hyperref}

\usepackage{mathtools}
\usepackage[english]{babel}
\usepackage{siunitx}
\usepackage[export]{adjustbox}

\usepackage{microtype}

\usepackage{booktabs}
\usepackage{makecell}
\usepackage{multirow}

\definecolor{pltred}{rgb}{0.839, 0.153, 0.157}

\DeclareMathOperator*{\argmin}{argmin}

\DeclareMathOperator*{\maximize}{maximize}

\DeclareMathOperator{\diag}{\mathrm{diag}}
\DeclareMathOperator{\tr}{\mathrm{tr}}

\newcommand{\R}{\mathbb{R}}

\newcommand{\ubar}[1]{\underaccent{\bar}{#1}}

\newcommand{\videourl}{http://tiny.cc/upright-robust}
\newcommand{\codeurl}{https://github.com/utiasDSL/upright}

\title{Robust Nonprehensile Object Transportation with Uncertain Inertial Parameters}
\author{Adam Heins and Angela P. Schoellig%
\thanks{This article was published in IEEE Robotics and Automation Letters. Digital Object Identifier (DOI): 10.1109/LRA.2025.3551067}%
\thanks{This work was supported by the Natural Sciences and Engineering Research Council of Canada and the Canadian Institute for Advanced Research.}%
  \thanks{The authors are with the Learning Systems and Robotics Lab (www.learnsyslab.org) at the Technical University of Munich, Germany, and the University of Toronto Institute for Aerospace Studies, Canada. They are also affiliated with the University of Toronto Robotics Institute, the Munich Institute of Robotics and Machine Intelligence (MIRMI), and the Vector Institute for Artificial Intelligence (e-mail: adam.heins@robotics.utias.utoronto.ca; angela.schoellig@tum.de).}%
}

\begin{document}

\maketitle

\begin{abstract}
  We consider the nonprehensile object transportation task known as the
  \emph{waiter's problem}---in which a robot must move an object on a
  tray from one location to another---when the transported object has uncertain
  inertial parameters. In contrast to existing approaches that completely
  ignore uncertainty in the inertia matrix or which only consider small
  parameter errors, we are interested in pushing the limits of the amount of
  inertial parameter uncertainty that can be handled. We first show how
  constraints that are robust to inertial parameter uncertainty can be
  incorporated into an optimization-based motion planning framework to transport objects while moving
  quickly. Next, we develop necessary conditions for the inertial parameters to
  be realizable on a bounding shape based on moment relaxations, allowing
  us to verify whether a trajectory will violate the constraints for
  \emph{any} realizable inertial parameters. Finally, we demonstrate our
  approach on a mobile manipulator in simulations and real hardware
  experiments: our proposed robust constraints consistently successfully transport
  a~56~cm tall object with substantial inertial parameter uncertainty in
  the real world, while the baseline approaches drop the object while
  transporting it.
\end{abstract}

\section{Introduction}

\IEEEPARstart{T}{he} \emph{waiter's problem}~\cite{flores2013time} is a nonprehensile
manipulation task that requires a robot to transport objects from one location
to another on a tray at the end effector (EE), like
a restaurant waiter. This manipulation task is called
\emph{nonprehensile}~\cite{lynch1996nonprehensile} because the objects are not
rigidly grasped: they are only attached to the robot by frictional contact and
thus retain some independent degrees of freedom (DOFs). Other examples of
nonprehensile manipulation include pushing, rolling, and
throwing~\cite{heins2024force,ruggiero2018nonprehensile}. A nonprehensile approach avoids
grasping and ungrasping phases and can handle delicate or unwieldy objects
which cannot be adequately grasped at all~\cite{pham2017admissible}; such an
approach for transporting objects is useful in industries including food service,
warehouse fulfillment, and manufacturing.

\begin{figure}[t]
  \centering
  \fbox{\includegraphics[width=\columnwidth]{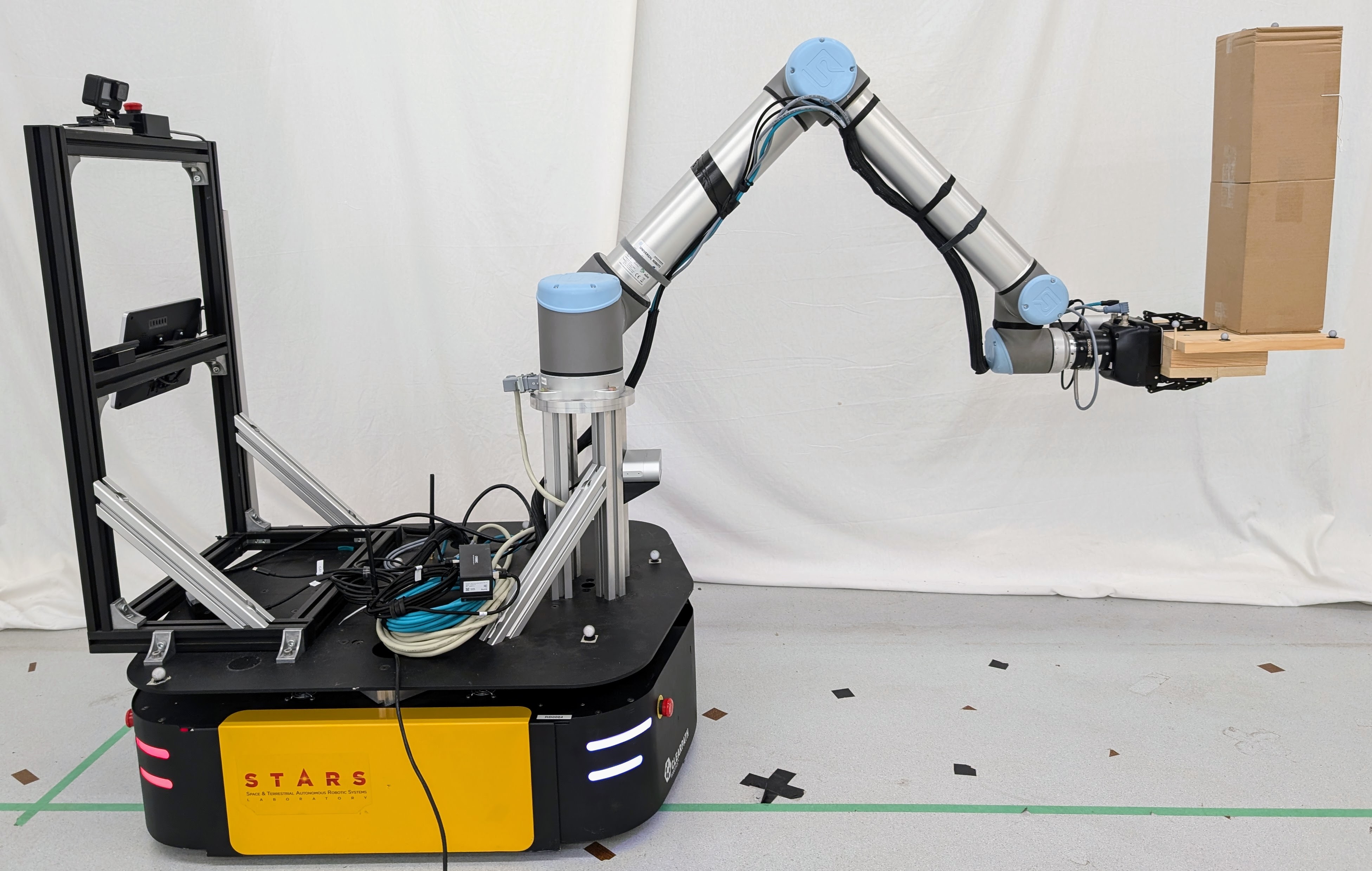}}
  \caption{The goal of this work is to move an object on a tray
  to a desired position without dropping it, despite the inertial parameters
  of the object being uncertain. Here our mobile manipulator is
  transporting a tall box with uncertain contents. A video of our
  experiments is available at \textbf{\texttt{\scriptsize \videourl}}.}
  \label{fig:eyecandy}
  \vspace{-10pt}
\end{figure}

We build on our previous work on the waiter's problem for mobile
manipulators~\cite{heins2023keep}. In contrast to~\cite{heins2023keep}, which
focused on fast online replanning to react to dynamic obstacles while balancing
objects with \emph{known} properties, here we focus on offline planning for a
transported object with \emph{unknown} inertial parameters---that is, the values
of the mass, center of mass (CoM), and inertia matrix are not known exactly but
rather lie in some set. Our approach is to plan trajectories to reach a desired
EE position while satisfying constraints that ensure the transported object does
not move with respect to the tray (see Fig.~\ref{fig:eyecandy}). These
\emph{sticking constraints} (so-called because they ensure the object ``sticks'' to the tray) depend on the geometric, frictional, and inertial
properties of the object. The geometry of the object can in principle be
estimated visually (e.g., using a camera), while frictional uncertainty can be
reduced by using a high-friction material for the tray surface or by using a
low friction coefficient in the planner~\cite{heins2023keep}. However, the
inertial properties can only be identified by moving the object around (see
e.g.~\cite{traversaro2016identification,wensing2018linear}), which is
time-consuming and could result in the object being dropped and damaged.
Instead, we propose using \emph{robust} constraints that successfully transport
the object despite the presence of substantial inertial parameter uncertainty.
Notably, we assume the CoM can be located at any height within the object, and
that the inertia matrix can take any physically realizable value.

We focus on balancing a single tall object with known geometry but
unknown mass and inertia matrix, and where the CoM is assumed to lie in a known
polyhedral convex set.
We use the object's known geometry to constrain the set of possible
inertial parameters. A set of inertial parameters can only be physically
realized on a given shape if there exists a corresponding mass density function
which is zero everywhere outside that shape~\cite{wensing2018linear}. We
develop necessary conditions for the inertial parameters to be physically
realizable on a bounding shape based on moment
relaxations~\cite{lasserre2009moments}. These realizability conditions allow us
to verify that a planned trajectory does not violate the sticking constraints
for \emph{any} physically realizable value of the inertial parameters,
providing theoretical guarantees for the robustness of our planned
trajectories. In summary, the contributions of this work are:
\begin{itemize}
  \item a planner for nonprehensile object transportation that explicitly
    handles objects with uncertain CoMs, extending the framework
    from~\cite{heins2023keep};
  \item a theoretical analysis of the sticking constraint satisfaction in the
    presence of a bounded CoM and any physically realizable inertia matrix,
    based on moment relaxations~\cite{lasserre2009moments};
  \item simulations and hardware experiments showing that our proposed
    robust constraints successfully transport the object---despite using tall
    objects with high inertial parameter uncertainty---while baseline
    approaches drop it;  
  \item an open-source implementation of our planner, available at
    \texttt{\small \codeurl}.
\end{itemize}

\section{Related Work}

The waiter's problem has been approached using a variety of methods including
offline
planning~\cite{pham2017admissible,zhou2022topp,gattringer2023point,brei2024serving},
online planning (i.e., model predictive
control)~\cite{selvaggio2023non,heins2023keep}, and reactive
control~\cite{moriello2018manipulating,muchacho2022a,selvaggio2022a,subburaman2023a}.
Few of these approaches address inertial parameter uncertainty in the transported
objects. One possible approach is to simulate the motion of a pendulum with the
EE, which naturally minimizes lateral forces acting on the transported object without
explicitly modelling it. However, so far this approach has only been used to
minimize slosh when transporting liquids~\cite{moriello2018manipulating,muchacho2022a}
rather than (uncertain) rigid bodies. Another approach
is~\cite{brei2024serving}, which develops a robust planner for the
waiter's problem based on reachability analysis, with parameter uncertainty
represented using polynomial zonotopes. In contrast to our
work,~\cite{brei2024serving} focuses on small amounts of uncertainty (e.g., a
5\% mass and inertia reduction) in both the transported objects and the links of
the robot. The resulting trajectories are also quite slow, with negligible
inertial acceleration (i.e., quasistatic). Instead, we achieve fast and dynamic
motion with tall objects under high parameter uncertainty (i.e., CoMs located
at any height in the object and \emph{any} realizable inertia matrix), but we
assume that uncertainty in the robot model is negligible (i.e., we use a
well-calibrated industrial robot).

Our formulation of robust sticking constraints draws inspiration from legged
robot balance
control~\cite{caron2015leveraging,caron2016multi,giftsun2017robustness,jiang2023locomotion}.
Indeed, the task of balancing an object on a tray is quite similar to balancing
a legged robot represented with a reduced-order model, which uses the
centroidal dynamics of the robot (i.e., the dynamics of a rigid body).
In~\cite{caron2016multi,giftsun2017robustness,jiang2023locomotion}, the CoM of
the robot is assumed to be uncertain and motions are generated that keep the
robot balanced for any possible CoM value in a polyhedral set. In particular,
we follow a similar approach to~\cite{jiang2023locomotion} for handling
polyhedral CoM uncertainty by enforcing sticking constraints corresponding to
a CoM located at each vertex of the set, and demonstrate its effectiveness for
the waiter's problem.
In contrast to these approaches, however, we are also interested in modelling
and handling uncertainty in the inertia matrix. While the CoM can reasonably
and intuitively be assumed to live in some convex polyhedral set, the inertia
matrix is more complicated. When considering the inertia matrix, the set of
physically realizable inertial parameters is \emph{spectrahedral} (i.e., it can
be represented using linear matrix inequalities (LMIs)) rather than polyhedral,
as discussed in~\cite{wensing2018linear}. We develop necessary conditions for
the inertial parameters to be physically realizable on a bounding shape, based
on moment relaxations~\cite{lasserre2009moments}, which we use as constraints
in a semidefinite program (SDP) to verify that our planned trajectories do not
violate any sticking constraints despite inertial parameter uncertainty.
Moment relaxations (i.e., Lasserre's hierarchy) have previously been applied in
robotics for tasks like certifiable localization~\cite{yang2023certifiably} and
trajectory planning~\cite{kang2024fast}, but not to bounds on the inertial
parameters of a rigid body.

\section{Background}

We begin with mathematical notation and preliminaries.

\subsection{Notation}

We denote the set of real numbers as~$\R$, the non-negative reals as~$\R_+$,
the non-negative integers as~$\mathbb{N}$, the~$n\times n$ symmetric matrices
as~$\mathbb{S}^n$, the symmetric positive semidefinite matrices
as~$\mathbb{S}^n_+$, and the symmetric positive definite matrices
as~$\mathbb{S}^n_{++}$. The notation~$\bm{A}\preccurlyeq\bm{B}$ means
that~$\bm{B}-\bm{A}\in\mathbb{S}^n_+$, while~$\leq$ denotes entry-wise
inequality. The~$n\times n$ identity matrix is denoted~$\bm{1}_n$. Finally, we
use~$\diag(\cdot)$ to construct (block) diagonal matrices.

\subsection{Polyhedron Double Description}

Any convex polyhedron~$\mathcal{P}$ can be described as either the convex hull of its
vertices or as a finite intersection of half spaces; this is called the
\emph{double description} of~$\mathcal{P}$. When~$\mathcal{P}$
is also a cone, it is called a polyhedral convex cone (PCC) and can be
described using either the \emph{face} or \emph{span} form~\cite{caron2015leveraging}:
\begin{equation*}
  \begin{aligned}
    \mathcal{P} &= \mathrm{face}(\bm{U}) = \{\bm{y} \mid \bm{U}\bm{y} \leq \bm{0} \} \\
                &= \mathrm{span}(\bm{V}) = \{\bm{V}\bm{z} \mid \bm{z} \geq \bm{0} \}.
  \end{aligned}
\end{equation*}
Following~\cite{balkcom2002computing}, we use the superscript~$(\cdot)^F$ to
denote conversion from span to face form, such
that~$\mathrm{face}(\bm{U}^F)=\mathrm{span}(\bm{U})$, and we use~$(\cdot)^S$ to
denote the conversion from face to span form. We perform the
conversions using the \texttt{cdd} library~\cite{fukuda1996double}.

\subsection{Moment Relaxations}\label{sec:moment}

The \emph{moment problem} asks when a sequence corresponds to the moments of
some Borel measure. We will make use of SDP relaxations for the
moment problem (\emph{moment relaxations}), which we briefly summarize here
from~\cite{lasserre2009moments}.
Define~$\mathbb{N}^n_d=\{\bm{\alpha}\in\mathbb{N}^n\mid \sum_{i=1}^n\alpha_i\leq d\}$.
Let~$\bm{r}\in\R^n$ be a point and let~$\bm{\alpha}\in\mathbb{N}^n_d$ be a vector of
exponents applied elementwise, such
that~$\bm{r}^{\bm{\alpha}}=r_1^{\alpha_1}\dots r_n^{\alpha_n}$.
Let~$f:\R^n\to\R$ be a polynomial of degree at most~$d$. Then we can write~$f(\bm{r}) = \sum_{\bm{\alpha}\in\mathbb{N}^n_d} f_{\bm{\alpha}}\bm{r}^{\bm{\alpha}} = \bm{f}^T\bm{b}_d(\bm{r})$,
where~$\bm{f}=\{f_{\bm{\alpha}}\}\in\R^{s(d)}$ is the vector of the
polynomial's coefficients with size~$s(d)\triangleq{n+d\choose d}$ and~$\bm{b}_d(\bm{r}) = [1, r_1,\dots,r_n,r_1^2,r_1r_2,\dots,r_n^d] \in \R^{s(d)}$
is the basis vector for polynomials of degree at most~$d$ in graded
lexicographical order. Given a
vector~$\bm{z}=\{z_{\bm{\alpha}}\}\in\R^{s(d)}$, the \emph{Riesz functional}
associated with~$\bm{z}$ is~$L_{\bm{z}}(f) =
\sum_{\bm{\alpha}\in\mathbb{N}^n_d}f_{\bm{\alpha}}z_{\bm{\alpha}}$, which maps
a polynomial~$f:\R^n\to\R$ of degree~$d$ to a scalar value. Given
a vector~$\bm{y}\in\R^{s(2d)}$,
the $d$th-order
\emph{moment matrix} associated with~$\bm{y}$ is~$\bm{M}_d(\bm{y}) = L_{\bm{y}}(\bm{b}_d(\bm{r})\bm{b}_d(\bm{r})^T) \in \R^{s(d)\times s(d)}$,
where~$L_{\bm{y}}$ is applied elementwise to each element of the
matrix~$\bm{b}_d(\bm{r})\bm{b}_d(\bm{r})^T$, such that the
element~$\bm{r}^{\bm{\alpha}}\bm{r}^{\bm{\beta}}=\bm{r}^{\bm{\alpha}+\bm{\beta}}$,
 with~$\bm{\alpha},\bm{\beta}\in\mathbb{N}^n_d$, is mapped to the
value~$y_{\bm{\alpha}+\bm{\beta}}$.
In addition, given a polynomial~$p:\R^n\to\R$ of degree~$2d$,
the \emph{localizing matrix} associated with~$p$ and~$\bm{y}$
is~$\bm{M}_d(p\bm{y})$. 

Suppose we want to determine if a given sequence~$\bm{y}\in\R^{s(2d)}$, known as
a \emph{truncated moment sequence (TMS)}, represents the moments of some Borel
measure~$\gamma$ supported in a compact semialgebraic
set~$\mathcal{K}=\{\bm{r}\in\R^n\mid p_j(\bm{r})\geq0,j=1,\dots,n_p\}$, where
each~$p_j(\bm{r})$ is a polynomial with degree~$2v_j$ (even) or~$2v_j-1$ (odd).
That is, we want to know if there exists~$\gamma:\R^n\to\R_+$ such that
\begin{equation*}
  \bm{M}_d(\bm{y}) = \int_\mathcal{K} \bm{b}_d(\bm{r})\bm{b}_d(\bm{r})^T\,d\gamma(\bm{r}),
\end{equation*}
which is known as the \emph{truncated K-moment problem (TKMP)}.
Define~$p_0(\bm{r})=1$ with~$v_0=0$.
Then a necessary condition for~$\gamma$ to exist (see Theorem 3.8
of~\cite{lasserre2009moments}) is
that for any~$r\geq d$, we can extend~$\bm{y}\in\R^{s(2d)}$
to~$\tilde{\bm{y}}\in\R^{s(2r)}$ while satisfying
\begin{equation}\label{eq:moment_necessary_conditions}
  \bm{M}_{r-v_j}(p_j\tilde{\bm{y}}) \succcurlyeq \bm{0}, \quad j=0,\dots,n_p.
\end{equation}
The moment constraints~\eqref{eq:moment_necessary_conditions} become tighter
as~$r$ increases, forming a hierarchy of SDP relaxations for the TKMP.

\section{Modelling}

Next, we present the models of the robot and object.

\subsection{Robot Model}

As in~\cite{heins2023keep}, we consider a velocity-controlled mobile
manipulator with
state~$\bm{x}=[\bm{q}^T,\bm{\nu}^T,\dot{\bm{\nu}}^T]^T\in\R^{n_x}$,
where~$\bm{q}$ is the generalized position, which includes the planar pose of
the mobile base and the arm's joint angles, and~$\bm{\nu}$ is the generalized
velocity. The input~$\bm{u}\in\R^{n_u}$ is the generalized jerk. We use a
kinematic model, which we represent generically as~$\dot{\bm{x}} =
\bm{a}(\bm{x}) + \bm{B}(\bm{x})\bm{u}$,
with~$\bm{a}(\bm{x})\in\R^{n_x}$ and~$\bm{B}(\bm{x})\in\R^{n_x\times n_u}$.
Though the actual commands sent to the robot are velocities, including
acceleration and jerk in the model allows us to reason about the sticking
constraints and encourage smoothness.

\subsection{Object Model}

We model the transported object as a rigid body subject to the
Newton-Euler equations
\begin{equation}\label{eq:obj_dynamics}
  \bm{w}_{\mathrm{C}} = \bm{w}_{\mathrm{GI}},
\end{equation}
where~$\bm{w}_{\mathrm{C}}$ is the contact wrench (CW)
and~$\bm{w}_{\mathrm{GI}}$ is the gravito-inertial wrench (GIW). All
quantities are expressed in a frame attached to the EE. We have
\begin{equation}\label{eq:giw}  
  \bm{w}_{\mathrm{GI}} \triangleq \bm{\Xi}\bm{\eta} - \mathrm{ad}(\bm{\xi})^T\bm{\Xi}\bm{\xi},
\end{equation}
where~$\bm{\Xi}\in\mathbb{S}^6_+$ is the object's spatial mass matrix,
$\bm{\xi}=[\bm{\omega}^T,\bm{v}^T]^T$ is the spatial velocity with angular
component~$\bm{\omega}\in\R^3$ and linear component~$\bm{v}\in\R^3$,
\begin{equation*}
  \mathrm{ad}(\bm{\xi}) \triangleq \begin{bmatrix}
    \bm{\omega}^\times & \bm{0} \\ \bm{v}^\times & \bm{\omega}^{\times}
  \end{bmatrix}
\end{equation*}
is the adjoint of~$\bm{\xi}$ with~$(\cdot)^{\times}$ forming a skew-symmetric matrix such
that~$\bm{a}^{\times}\bm{b}=\bm{a}\times\bm{b}$ for any~$\bm{a},\bm{b}\in\R^3$,
and~$\bm{\eta}=[\dot{\bm{\omega}}^T,\dot{\bm{v}}^T-\bm{g}^T]^T$
with gravitational acceleration~$\bm{g}\in\R^3$. The mass matrix
is defined as
\begin{equation*}
  \bm{\Xi} \triangleq \begin{bmatrix}
    \bm{I} & m\bm{c}^\times \\
    -m\bm{c}^\times & m\bm{1}_3
  \end{bmatrix},
\end{equation*}
where~$m$ is the object's mass, $\bm{c}$ is the position of the object's CoM,
and~$\bm{I}$ is its inertia matrix.

\section{Robust Sticking Constraints}\label{sec:robust}

We want to enforce constraints on the robot's motion so that the transported
object does not move relative to the EE (i.e., it ``sticks'' to the EE). These
sticking constraints prevent the object from sliding, tipping, or breaking
contact. This is known as a \emph{dynamic grasp}~\cite{mason1993dynamic}.

\subsection{Contact Force Constraints}

\begin{figure}[t]
  \centering
    \includegraphics[width=0.75\columnwidth]{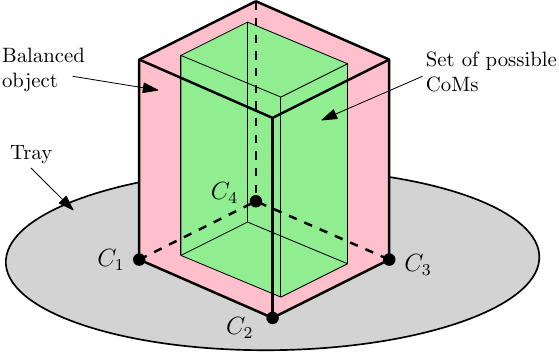}
    \caption{A box (red) on a tray, with four contact
    points~$C_1$--$C_4$ located at the vertices of the base. We assume that the
    box's center of mass (CoM) is not known exactly, but rather only known to
    lie inside some polyhedral set (green).}
  \label{fig:3d_diagram}
  \vspace{-10pt}
\end{figure}

We can ensure an object sticks to the EE by including all contact forces
directly into the motion planner and constraining the solution to be consistent
with the desired (sticking) dynamics. Suppose there are~$n_c$ contact
points~$\{C_i\}_{i=1}^{n_c}$ between the object and tray (see
Fig.~\ref{fig:3d_diagram}), with corresponding contact
forces~$\{\bm{f}_i\}_{i=1}^{n_c}$. By Coulomb's law, each contact
force~$\bm{f}_i\in\R^3$ must lie inside its friction cone, which we linearize
to obtain the set of constraints~$\bm{F}_i\bm{f}_i\leq\bm{0}$, where
\begin{equation*}
  \bm{F}_i = \begin{bmatrix}
    0 & 0 & -1 \\
    1 & 1 & -\mu_i \\ 1 & -1 & -\mu_i \\ -1 & 1 & -\mu_i \\ -1 & -1 & -\mu_i
  \end{bmatrix}\begin{bmatrix} \hat{\bm{t}}_{i_1} & \hat{\bm{t}}_{i_2} & \hat{\bm{n}}_i \end{bmatrix}^T,
\end{equation*}
with friction coefficient~$\mu_i$, contact normal~$\hat{\bm{n}}_i$, and
orthogonal contact tangent directions~$\hat{\bm{t}}_{i_1}$ and~$\hat{\bm{t}}_{i_2}$.
Letting~$\bm{\zeta}=[\bm{f}_1^T,\dots,\bm{f}_{n_c}^T]^T$, we can
write the linearized friction cone constraints for all~$n_c$ contact forces in
matrix form as
\begin{equation}\label{eq:friction_cone_all}
  \bm{F}\bm{\zeta} \leq \bm{0},
\end{equation}
where~$\bm{F}=\diag(\bm{F}_1, \dots,\bm{F}_{n_c})$. The CW on the object is
\begin{equation}\label{eq:contact_wrench}
  \bm{w}_{\mathrm{C}} \triangleq \begin{bmatrix} \bm{\tau}_\mathrm{C} \\ \bm{f}_\mathrm{C} \end{bmatrix} = \sum_{i=1}^{n_c}\bm{G}_i\bm{f}_i,
\end{equation}
where~$\bm{\tau}_{\mathrm{C}}$ and~$\bm{f}_{\mathrm{C}}$ are the total contact
torque and force and~$\bm{G}_i = [-\bm{r}_i^\times, \bm{1}_3]^T$
is the~$i$th contact Jacobian with~$\bm{r}_i$ the location of~$C_i$.
The object sticks to the EE at a given time instant if a set of
contact forces can be found
satisfying~\eqref{eq:obj_dynamics}--\eqref{eq:contact_wrench}. This approach
can handle any number of (possibly non-coplanar) contact points, with
surface contacts represented as polygons with a contact point at each vertex.

The set~$\mathcal{W}_{\mathrm{C}}$ of all possible contact wrenches is known as
the \emph{contact wrench cone} (CWC)~\cite{caron2015leveraging}, which is a PCC
containing all CWs that can be produced by feasible contact forces. We
have~$\mathcal{W}_{\mathrm{C}} = \{\bm{G}\bm{\zeta} \mid
\bm{F}\bm{\zeta}\leq\bm{0} \}$, where~$\bm{G}=[\bm{G}_1,\dots,\bm{G}_{n_c}]$ is
known as the \emph{grasp matrix}. From~\eqref{eq:obj_dynamics}, we
must have~$\bm{w}_{\mathrm{GI}}\in\mathcal{W}_{\mathrm{C}}$ for the object to
remain stationary relative to the EE.

\subsection{Robustness to Inertial Parameter Uncertainty}

Let~$\bm{\theta}=[m,m\bm{c}^T,\mathrm{vech}(\bm{I})^T]^T\in\R^{10}$ be the
inertial parameter vector for the object,
where~$\mathrm{vech}(\bm{I})\in\R^6$ is the \emph{half-vectorization}
of~$\bm{I}$~\cite{traversaro2016identification}. Furthermore, let us assume
that the exact value of~$\bm{\theta}$ is unknown, but that it lies
inside a set~$\Theta$. Since~$\bm{w}_{\mathrm{GI}}$ is linear
in~$\bm{\theta}$~\cite{traversaro2016identification}, we can write the set of
possible GIWs under inertial parameter uncertainty as
\begin{equation*}
  \mathcal{W}_{\mathrm{GI}}(\bm{\xi},\bm{\eta}) = \{\bm{Y}(\bm{\xi},\bm{\eta})\bm{\theta}\mid \bm{\theta}\in\Theta \}
\end{equation*}
where~$\bm{Y}(\bm{\xi},\bm{\eta})\in\R^{6\times10}$ is known as the
\emph{regressor matrix}. To ensure the sticking constraints are
satisfied for any~$\bm{\theta}\in\Theta$, we want to generate EE
motions~$(\bm{\xi},\bm{\eta})$ that satisfy
\begin{equation}\label{eq:WGI_in_WC}
  \mathcal{W}_{\mathrm{GI}}(\bm{\xi},\bm{\eta})\subseteq\mathcal{W}_{\mathrm{C}}
\end{equation}
at all timesteps. More concretely, \eqref{eq:WGI_in_WC} is satisfied if and only
if, for each~$\bm{\theta}\in\Theta$, there exists a set of contact
forces~$\bm{\zeta}$
satisfying~$\bm{Y}(\bm{\xi},\bm{\eta})\bm{\theta}=\bm{G}\bm{\zeta}$
and~$\bm{F}\bm{\zeta}\leq\bm{0}$.

We can
enforce~\eqref{eq:WGI_in_WC} using only constraints on the extreme
points~$\mathrm{ex}(\Theta)$ of~$\Theta$. To see this, observe that
since~$\mathcal{W}_{\mathrm{C}}$ is convex, any convex combination of points
in~$\mathcal{W}_{\mathrm{C}}$ also lies in~$\mathcal{W}_{\mathrm{C}}$. And
since any~$\bm{\theta}\in\Theta$ is a convex combination of points
in~$\mathrm{ex}(\Theta)$, it follows
that~$\bm{Y}(\bm{\xi},\bm{\eta})\bm{\theta}\in\mathcal{W}_{\mathrm{C}}$ for
any~$\bm{\theta}\in\Theta$ as long
as~$\bm{Y}(\bm{\xi},\bm{\eta})\bm{\theta}\in\mathcal{W}_{\mathrm{C}}$ for
all~$\bm{\theta}\in\mathrm{ex}(\Theta)$.

Furthermore, for lightweight objects, we can ignore the value of the object's mass when transporting a single object.\footnote{We are assuming that the object has a small enough mass for the robot's internal position controller to accurately track the desired trajectory despite the uncertain payload.} To see this, suppose the true inertial parameter vector
is~$\bm{\theta}=[m,m\bm{c}^T,\mathrm{vech}(\bm{I})^T]^T\in\Theta$.
Since~$\mathcal{W}_{\mathrm{C}}$ is a convex cone and~$m>0$, it
follows that~$\bm{Y}(\bm{\xi},\bm{\eta})\bm{\theta}\in\mathcal{W}_{\mathrm{C}}$ if and
only if~$\bm{Y}(\bm{\xi},\bm{\eta})\skew3\hat{\bm{\theta}}\in\mathcal{W}_{\mathrm{C}}$,
where~$\skew3\hat{\bm{\theta}}\triangleq\bm{\theta}/m$
is the \emph{mass-normalized} parameter vector.
In other words, for any~$\bm{\theta}\in\Theta$, we can always
instead use~$\skew3\hat{\bm{\theta}}$ to enforce the sticking constraints,
which is independent of the true mass. This result holds for any object that is
not in contact with any others (except of course the tray). However, when
multiple objects are in contact with each other, the force transmitted between them
depends on their relative masses, so we can no longer ignore them.

Finally, we will also ignore uncertainty in the inertia matrix \emph{while
planning} trajectories, as it would be expensive to enforce the LMI
constraint required for physical realizability~\cite{wensing2018linear}.
Instead, we will check our trajectories \emph{after planning} to verify that
the sticking constraints are satisfied for any physically realizable inertial
parameters---our experiments suggest that handling large uncertainty in the CoM
gives us enough robustness to handle any physically realizable inertia matrix
as well. We will return to the analysis of physically realizable inertial
parameters in Sec.~\ref{sec:verify}. For now, since we are ignoring uncertainty
in the mass and inertia matrix, we are left with uncertainty in the CoM.
Similar to~\cite{jiang2023locomotion}, our approach therefore is to assume
that~$\bm{c}$ belongs to a known polyhedral set~$\mathcal{C}$ with~$n_v$
vertices (see Fig.~\ref{fig:3d_diagram}) and enforcing sticking constraints
for~$n_v$ objects, which differ only in that the~$i$th object has CoM located
at the~$i$th vertex of the CoM uncertainty set. This is equivalent to
enforcing~\eqref{eq:WGI_in_WC}.

\section{Robust Planning}

Our goal is to generate a state-input trajectory for our robot that satisfies
the robust sticking constraints developed in the previous section. We
formulate the motion planning problem as a constrained optimal control problem
(OCP), similar to the formulation in~\cite{heins2023keep}. In contrast
to~\cite{heins2023keep}, however, which solves the OCP online in a model
predictive control framework, we solve it once offline with a longer time
horizon and then track the resulting optimal trajectory; we found this approach
to be more reliable in our experiments. In particular, we found that a
longer time horizon converged to the desired position faster with fewer
oscillations, and that updating the desired trajectory online could cause
additional EE vibration that made it more likely for tall, uncertain
objects to be dropped.
The trajectories~$\bm{x}(t)$, $\bm{u}(t)$,
and~$\bm{\zeta}(t)$ are optimized over a duration~$T$ by solving a nonlinear
optimization problem. Suppressing the time dependencies, the problem is
\begin{equation}\label{eq:traj_opt_prob}
  \begin{aligned}
    \argmin_{\bm{x},\bm{u},\bm{\zeta}} &\quad \frac{1}{2}\int_{0}^{T} \ell(\bm{x},\bm{u})\,dt \\
    \text{subject to} &\quad \dot{\bm{x}} = \bm{a}(\bm{x})+\bm{B}(\bm{x})\bm{u} & \text{(robot model)} \\
                      &\quad \mathcal{W}_{\mathrm{GI}}(\bm{x})\subseteq\mathcal{W}_{\mathrm{C}} & \text{(sticking)} \\
                      &\quad \ubar{\bm{x}} \leq \bm{x} \leq \bar{\bm{x}} &\text{(state limits)} \\
                      &\quad \ubar{\bm{u}} \leq \bm{u} \leq \bar{\bm{u}} &\text{(input limits)} \\
                      &\quad \bm{\varphi}(\bm{x}_f) = \bm{0} &\text{(terminal)}
  \end{aligned}
\end{equation}
where the stage cost is
\begin{equation*}
  \ell(\bm{x},\bm{u}) = \|\Delta\bm{r}(\bm{x})\|^2_{\bm{W}_r} + \|\bm{x}\|^2_{\bm{W}_x} + \|\bm{u}\|^2_{\bm{W}_u},
\end{equation*}
with~$\|\cdot\|^2_{\bm{W}}=(\cdot)^T\bm{W}(\cdot)$ for weight matrix~$\bm{W}$,
and~$\Delta\bm{r}(\bm{x})=\bm{r}_{d}-\bm{r}_{e}(\bm{x})$ is the EE position
error between the desired position~$\bm{r}_d$ and the current
position~$\bm{r}_e(\bm{x})$. The sticking constraints implicitly depend on the
contact forces~$\bm{\zeta}$, and we have expressed~$\mathcal{W}_{\mathrm{GI}}$
as a function of~$\bm{x}$ via forward kinematics. The terminal
constraint~$\bm{\varphi}(\bm{x}_f) =
[\Delta\bm{r}(\bm{x}_f)^T,\bm{\nu}_f^T,\dot{\bm{\nu}}_f^T]^T=\bm{0}$ acts only on
the final state~$\bm{x}_f\triangleq\bm{x}(T)$ and steers it toward a
stationary state with no position error.

We solve~\eqref{eq:traj_opt_prob} by discretizing the planning
horizon~$T$ with a fixed timestep~$\Delta t$ and using sequential
quadratic programming (SQP) via the open-source framework OCS2~\cite{ocs2} and
quadratic programming solver HPIPM~\cite{frison2020hpipm}, with required Jacobians
computed using automatic differentiation. We use the Gauss-Newton
approximation for the Hessian of the cost and we soften all state
limits and sticking constraints:
given a generic state constraint~$g(\bm{x})\geq0$, we add a slack
variable~$s\geq0$ to relax the constraint to~$g(\bm{x})+s\geq0$, and the $L_2$
penalty~$w_ss^2$ is added to the cost with weight~$w_s>0$.
Like~\cite{heins2023keep}, we also plan
while assuming that there is \emph{zero} contact friction between the tray and
transported object (i.e., tangential forces are to be avoided, but small ones are allowed because the constraints are soft). This provides robustness to uncertain friction and other
disturbances while also reducing the computational cost of
solving~\eqref{eq:traj_opt_prob}, since each contact force variable
need only be represented by a single non-negative scalar representing the
normal force (see~\cite{heins2023keep} for more details). Given the
soft constraints and time discretization of~\eqref{eq:traj_opt_prob}, robust
sticking is not \emph{guaranteed} but is heavily \emph{encouraged.}

Once we have solved~\eqref{eq:traj_opt_prob} to obtain the planned
optimal
trajectory~$\bm{x}_d(t)=[\bm{q}_d^T(t),\bm{\nu}_d^T(t),\dot{\bm{\nu}}_d^T(t)]^T$,
we need to track it online. At each control timestep, we generate the commanded
velocity~$\bm{\nu}_{\mathrm{cmd}}$ using the simple affine control
law~$\bm{\nu}_{\mathrm{cmd}}=\bm{K}_p(\bm{q}_d-\bm{q})+\bm{\nu}_d$,
where~$\bm{K}_p\in\mathbb{S}^9_{++}$ is a gain matrix.

\section{Verifying Sticking Constraint Satisfaction}\label{sec:verify}

Let us now consider uncertainty in the inertia matrix. We want a way to show that our
choice to ignore inertia matrix uncertainty in the planner is justified; that
is, given a trajectory, we want to verify that the sticking constraints are
not violated for \emph{any} realizable value of the inertia matrix. In this
section we develop an SDP to determine an upper bound on the maximum constraint
violation given the uncertain inertial parameters and bounding shape for the
object.

\subsection{Double Description of the Contact Wrench Cone}\label{sec:dd}

Following~\cite{caron2015leveraging}, we build the face form of the CWC. First,
notice that~\eqref{eq:friction_cone_all} describes a
PCC~$\mathrm{face}(\bm{F})$. Converting~\eqref{eq:friction_cone_all} to span
form, we
have~$\mathcal{W}_{\mathrm{C}}=\{\bm{G}\bm{F}^S\bm{z}\mid\bm{z}\geq\bm{0}\}$.
Next, letting~$\bm{H}=(\bm{G}\bm{F}^S)^F\in\R^{n_h\times6}$, we have the face
form of the
CWC~$\mathcal{W}_{\mathrm{C}}=\{\bm{w}\in\R^6\mid\bm{H}\bm{w}\leq\bm{0}\}$.
This form allows us to write the robust sticking
constraints~\eqref{eq:WGI_in_WC} as
\begin{equation}\label{eq:balancing_constraint_equations}
  \bm{H}\bm{Y}(\bm{\xi},\bm{\eta})\bm{\theta} \leq \bm{0}\ \ \forall\bm{\theta}\in\Theta.
\end{equation}
Let~$\bm{h}_i^T$ be the~$i$th row of~$\bm{H}$. Then we can rewrite the
constraint~\eqref{eq:balancing_constraint_equations} as a set of inner
optimization problems
\begin{equation}\label{eq:inner_opt}
  \left(\maximize_{\bm{\theta}\in\Theta}\;\;\bm{h}_i^T\bm{Y}(\bm{\xi},\bm{\eta})\bm{\theta}\right) \leq 0,
\end{equation}
with one problem for each of the~$n_h$ rows. This inequality form of the
sticking constraints will allow us to solve for the worst-case value of each
constraint. However, we first need to determine appropriate bounds on the
set~$\Theta$.

\subsection{Worst-Case Sticking Constraints}\label{sec:worst-case}

Let us apply the moment relaxation machinery from Sec.~\ref{sec:moment} to the problem of physically realizable
inertial parameters. We have dimension~$n=3$ and degree~$d=1$. The TMS~$\bm{y}\in\R^{10}$ and
the associated moment matrix~$\bm{M}_1(\bm{y})$ can be used to represent the
inertial parameters of a rigid body, with the relationship
\begin{equation}\label{eq:permuted_pim}
  \bm{M}_1(\bm{y}) = \begin{bmatrix} m & m\bm{c}^T \\ m\bm{c} & \bm{S} \end{bmatrix},
\end{equation}
where~$\bm{S}=(1/2)\tr(\bm{I})\bm{1}_3-\bm{I}$~\cite{wensing2018linear}.\footnote{A permuted version of~\eqref{eq:permuted_pim}, known
as the \emph{pseudo-inertia matrix}, is the more common form in the robot
dynamics literature (see e.g.~\cite{wensing2018linear}), but here we maintain
consistency with the TKMP literature~\cite{lasserre2009moments}.} We want to constrain the inertial
parameters to correspond to a mass density~$\rho:\R^3\to\R_+$ supported
entirely in a compact bounding shape~$\mathcal{K}=\{\bm{r}\in\R^3 \mid
p_j(\bm{r})\geq0,j=1,\dots,n_p\}$ that contains the transported object,
where each~$p_j$ is a polynomial of degree~$2v_j$ or~$2v_j-1$, such that
\begin{equation*}
  \bm{M}_1(\bm{y}) = \int_{\mathcal{K}} \bm{b}_1(\bm{r})\bm{b}_1(\bm{r})^T\,d\rho(\bm{r}).
\end{equation*}
For simplicity, we assume our bounding shape is a convex polyhedron, so~$v_j=1$
for each~$j=1,\dots,n_p$, and we take~$p_0(\bm{r})=1$ with~$v_0=0$. Given
an EE trajectory, we want to determine if any realizable value of the inertial
parameters would violate the sticking constraints~\eqref{eq:inner_opt} at any
time. That is, we would like to know if the optimal value of
\begin{equation}\label{eq:max_constraint_prob}
  \maximize_{\bm{\theta}\in\Theta} \quad \bm{h}_i^T\bm{Y}(\bm{\xi},\bm{\eta})\bm{\theta}
\end{equation}
is positive for any row~$\bm{h}_i^T$ of~$\bm{H}$ at any time instant of the
trajectory. Using the conditions~\eqref{eq:moment_necessary_conditions} with
order~$r=2$, the problem~\eqref{eq:max_constraint_prob} can be
relaxed to the SDP
\begin{equation}\label{eq:max_constraint_sdp}
  \begin{aligned}
    \maximize_{\bm{\theta},\tilde{\bm{y}}} &\quad \bm{h}_i^T\bm{Y}(\bm{\xi},\bm{\eta})\bm{\theta} \\
    \text{subject to} &\quad \bm{M}_1(\tilde{\bm{y}}) = \begin{bmatrix} 1 & \bm{c}(\bm{\theta})^T \\ \bm{c}(\bm{\theta}) & \bm{S}(\bm{\theta}) \end{bmatrix}, \\
                      &\quad \bm{M}_{2-v_j}(p_j\tilde{\bm{y}}) \succcurlyeq \bm{0}, \quad j=0,\dots,n_p, \\
                      &\quad \bm{c}(\bm{\theta}) \in \mathcal{C},
  \end{aligned}
\end{equation}
where~$\tilde{\bm{y}}\in\R^{20}$ is an extended TMS, we have constrained
the CoM~$\bm{c}$ to be located within some convex
polyhedron~$\mathcal{C}\subset\mathcal{K}$, and we have fixed~$m=1$ since the
sticking constraints are independent of mass for a single object. We have also
made the dependencies on the decision variable~$\bm{\theta}$ explicit for
clarity; note that~$\bm{S}$ and~$\bm{c}$ depend linearly on~$\bm{\theta}$.

Since~\eqref{eq:max_constraint_sdp} is a relaxation
of~\eqref{eq:max_constraint_prob}, its optimal value is an upper bound on the
maximum possible violation for each constraint.
Notably,~\eqref{eq:max_constraint_sdp} accounts for all possible values of the
inertia matrix. We verify that a planned trajectory is robust to
inertial parameter uncertainty by
solving~\eqref{eq:max_constraint_sdp} pointwise at a fixed frequency
along the trajectory. If the optimal value of~\eqref{eq:max_constraint_sdp} is
always non-positive, then the constraint is not violated; we assume that the time discretization is fine enough that the constraints are not violated between timesteps. Furthermore, we can
verify robustness to uncertain friction coefficients at the same time by
constructing~$\mathcal{W}_{\mathrm{C}}$ and thus~$\bm{H}$ with low friction
coefficients: if the constraints are never violated, then any combination of
higher friction coefficients and realizable inertial parameters will also
satisfy the constraints.

One could also consider enforcing full realizability
constraints directly in the planning problem~\eqref{eq:traj_opt_prob}
to ensure the planned trajectories are robust a priori. However, this would be
computationally expensive and numerically challenging because of the LMI
constraints required for physical realizability combined with the nonlinearity
of the problem. We leave this for future work.

\section{Simulation Experiments}\label{sec:sim}

We begin the evaluation of our proposed robust planning approach in simulation
using the PyBullet simulator and a simulated version of our experimental
platform, a $9$-DOF mobile manipulator consisting of a Ridgeback mobile base
and UR10 arm, shown in Fig.~\ref{fig:eyecandy}. In all experiments
(simulated and real) we use~$\Delta t=\SI{0.1}{s}$, $T=\SI{10}{s}$, and weights
\begin{align*}
  \bm{W}_r &= \bm{1}_3, & \bm{W}_x &= \diag(0\bm{1}_9,10^{-1}\bm{1}_9,10^{-2}\bm{1}_9), \\
  \bm{W}_u &= 10^{-3}\bm{1}_9, & w_s &= 100.
\end{align*}
The state and input limits are
\begin{alignat*}{4}
  \bar{\bm{q}} &= \begin{bmatrix} 10\bm{e}_3 \\ 2\pi\bm{e}_6 \end{bmatrix}, &\;
  \bar{\bm{\nu}} &= \begin{bmatrix} 1.1\bm{e}_2 \\ 2\bm{e}_3 \\ 3\bm{e}_4 \end{bmatrix}, &\;
  \dot{\bar{\bm{\nu}}} &= \begin{bmatrix} 2.5\bm{e}_2 \\ 1 \\ 10\bm{e}_6 \end{bmatrix}, &\;
  \bar{\bm{u}} &= \begin{bmatrix} 20\bm{e}_3 \\ 80\bm{e}_6 \end{bmatrix},
\end{alignat*}
where~$\bar{\bm{x}}=[\bar{\bm{q}}^T,\bar{\bm{\nu}}^T,\dot{\bar{\bm{\nu}}}^T]^T$,
$\ubar{\bm{x}}=-\bar{\bm{x}}$, $\ubar{\bm{u}}=-\bar{\bm{u}}$, and~$\bm{e}_n$
denotes an $n$-dimensional vector of ones. The control gain
is~$\bm{K}_p=\bm{1}_9$ and the simulation timestep is~\SI{0.1}{ms}.
All experiments are run on a standard
laptop with eight Intel Xeon CPUs at~\SI{3}{GHz} and~\SI{16}{GB} of RAM.

\begin{figure}[t]
  \centering
    \includegraphics[width=0.75\columnwidth]{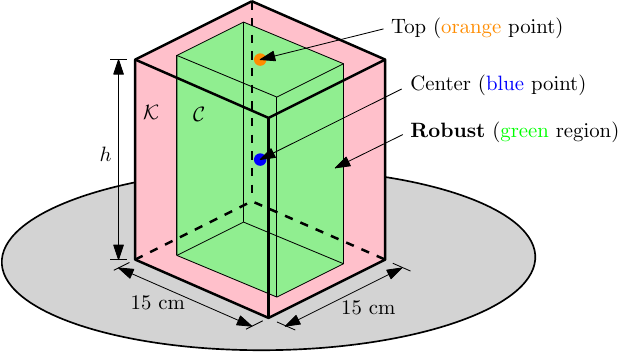}
    \caption{We transport a cuboid-shaped object~$\mathcal{K}$ (red)
      with an uncertain CoM contained in~$\mathcal{C}$ (green). We
      test three variations of the sticking constraints: assume the CoM is at
      the object's centroid (Center), assume it is centered at the top of the
      object (Top), and Robust, in which the controller enforces sticking constraints
      for eight different objects, where each has its CoM at one of the
      vertices of~$\mathcal{C}$.}
  \label{fig:sim_box}
\end{figure}

We are interested in cases where the inertial parameters of the transported object
are uncertain and this uncertainty can result in task failures (i.e., the
object is dropped) if the uncertainty is ignored. We use a tall
box~$\mathcal{K}$ with a $\SI{15}{cm}\times\SI{15}{cm}$ base and height~$h$
(see Fig.~\ref{fig:sim_box}) as the transported object. This object is tall
relative to its support area and therefore prone to tipping over, particularly
if the inertial parameters are not known exactly. Suppose we assume that the
CoM can lie anywhere in a box~$\mathcal{C}$ with dimensions
$\SI{12}{cm}\times\SI{12}{cm}\times h$, centered within~$\mathcal{K}$; that is,
the CoM can be located \emph{anywhere} in the object as long as it is at least
\SI{1.5}{cm} from the sides. The simulated friction coefficient between the box
and tray is~$\mu=0.2$. We compare three sets of sticking constraints
(again, see Fig.~\ref{fig:sim_box}):
\begin{itemize}
  \item \textbf{Center:} The CoM is located at the center of~$\mathcal{C}$;
  \item \textbf{Top:} The CoM is centered on the top face of~$\mathcal{C}$;
  \item \textbf{Robust:} A set of eight sticking constraints are used,
    corresponding to a CoM at each vertex of~$\mathcal{C}$.
\end{itemize}
In all cases, the inertia matrix used in the planner is set to correspond to a
uniform mass density. The first two constraint methods are baselines where we
are not explicitly accounting for the uncertainty in the parameters.
Intuitively, it is more difficult to transport an object with a higher CoM, so we
may expect the Top constraints to be more successful than the Center
constraints. In contrast to these baselines, our proposed Robust constraints
explicitly handle uncertainty in the CoM.

We test all combinations of the following: three
desired positions~$\bm{r}_{d_1}=[-2,1,0]^T$, $\bm{r}_{d_2}=[0,2,0.25]^T$,
and~$\bm{r}_{d_3}=[2,0,-0.25]^T$ (in meters), 15 different simulated CoM
positions (one at the center of~$\mathcal{C}$, eight at the vertices
of~$\mathcal{C}$, and six at the centers of the faces of~$\mathcal{C}$), and
three different values for the inertia matrix, computed as follows. We solve a
convex optimization problem to find the diagonal inertia matrix about the CoM
which corresponds to a set of point masses at the vertices of~$\mathcal{K}$
with the maximum smallest eigenvalue. This gives us a large realizable inertia
matrix value~$\bm{I}$; we then also test with the smaller inertia
values~$0.5\bm{I}$ and~$0.1\bm{I}$. The simulated mass is fixed
to~$m=\SI{1}{kg}$. We test each of the 135 total combinations of trajectory,
CoM, and inertia matrix for different object heights~$h$ and constraint
methods.

\begin{figure}[t]
  \centering
    \includegraphics[width=\columnwidth]{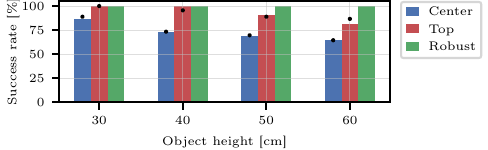}
    \caption{Success rate of the different types of sticking
    constraints for all 135 combinations of desired positions, CoMs, and inertias
    for each object height and constraint method (1620 total runs). The Robust
    constraints always successfully transport the object, while the other constraint types
    result in an increasing number of failures as the object height increases;
    the rate varies slightly with the number of SQP iterations~$n_s$.
    Here the bar shows~$n_s=3$ and the black dot~$n_s=10$.}
  \label{fig:success_rate_solve_time}
  \vspace{5pt}
\end{figure}

The success rates for the simulations are shown in
Fig.~\ref{fig:success_rate_solve_time}. The success rate is the percentage of
runs (out of the 135 total per object) that successfully deliver the object to
the goal without it being dropped from the tray. The Robust
constraints are always successful (with a maximum object displacement of
only~\SI{2}{mm}). The Top and Center constraints produce fewer
successful runs as the object height increases, which suggests that the sticking constraints are more sensitive to parameter error for taller objects. The success rate of the Top and Center constraints varies slightly when using more SQP iterations~$n_s$ to
solve~\eqref{eq:traj_opt_prob}, at the cost of a longer solve time. We report
the results for~$n_s=3$ and~$n_s=10$ (the Robust constraints are completely
successful for both values of~$n_s$, so we only report~$n_s=3$); increasing
to~$n_s=20$ did not produce a higher success rate. The average solve times for
the Center constraints are~\SI{110}{ms} ($n_s=3$) and~\SI{310}{ms} ($n_s=10$);
for the Top constraints~\SI{111}{ms} ($n_s=3$) and~\SI{324}{ms} ($n_s=10$); and
for the Robust constraints~\SI{324}{ms}. Even when more compute time is used to
refine the trajectory, the baselines are still not as successful as the Robust constraints.

While the results in~Fig.~\ref{fig:success_rate_solve_time} show that our
proposed constraints are robust for \emph{particular} combinations of CoM
positions and inertia matrices, Table~\ref{tab:realizability} shows the maximum
possible sticking constraint violations for \emph{any} realizable inertial
parameter value, obtained by solving~\eqref{eq:max_constraint_sdp} at each
point along the planned trajectory (discretized with a~\SI{10}{ms} interval).
Solving~\eqref{eq:max_constraint_sdp}
for all~$i=1,\dots,n_h$ at a single timestep took about~\SI{2.2}{s}.
The Robust constraints have no violation for any possible value of the inertia
matrix (while the Center and Top constraints always do), justifying our
decision to ignore uncertainty in the inertia matrix within the planner. While
negative violation implies failure should not occur; positive violation does
not mean that it \emph{will} occur.

\begin{table}[t]
  \caption{Maximum constraint violation for each object height and sticking
    constraint method using the moment conditions for physical realizability. A
    negative value means that the constraints are never violated for any
    realizable inertial parameters, which is only the case using our proposed
    Robust constraints. These results are for~$n_s=3$; the values
    with~$n_s=10$ are very similar.}
  \centering
  \begin{tabular}{c c c c}
    \toprule
    Height [cm] & Center & Top & Robust \\
    \midrule
    30 & $2.48$  & $4.51$  & $-1.15$ \\
    40 & $5.03$  & $7.19$  & $-0.97$ \\
    50 & $8.21$  & $9.09$  & $-0.77$ \\
    60 & $11.96$ & $10.50$ & $-0.56$ \\
    \bottomrule
  \end{tabular}
  \label{tab:realizability}
\end{table}

Finally, recall that the simulated value of the friction coefficient between
the tray and object is~$\mu=0.2$. This value does not impact the behavior of
the planner because the planner assumes~$\mu=0$ and then tries to satisfy the
softened sticking constraints approximately. However, the underlying value
of~$\mu$ can potentially affect the amount of constraint violation,
since~$\bm{h}_i$ in~\eqref{eq:max_constraint_sdp} depends on the friction
coefficient. Interestingly, we evaluated the constraint violation for the
robust constraints with~$h=\SI{60}{cm}$ and~$\mu=0.1$ and found the maximum
constraint violation to be the same as with~$\mu=0.2$, which suggests that the
friction coefficient is not the limiting factor when transporting tall objects like
those used here.

\section{Hardware Experiments}

\begin{figure}[t]
  \centering
  \fbox{\includegraphics[width=0.6\columnwidth]{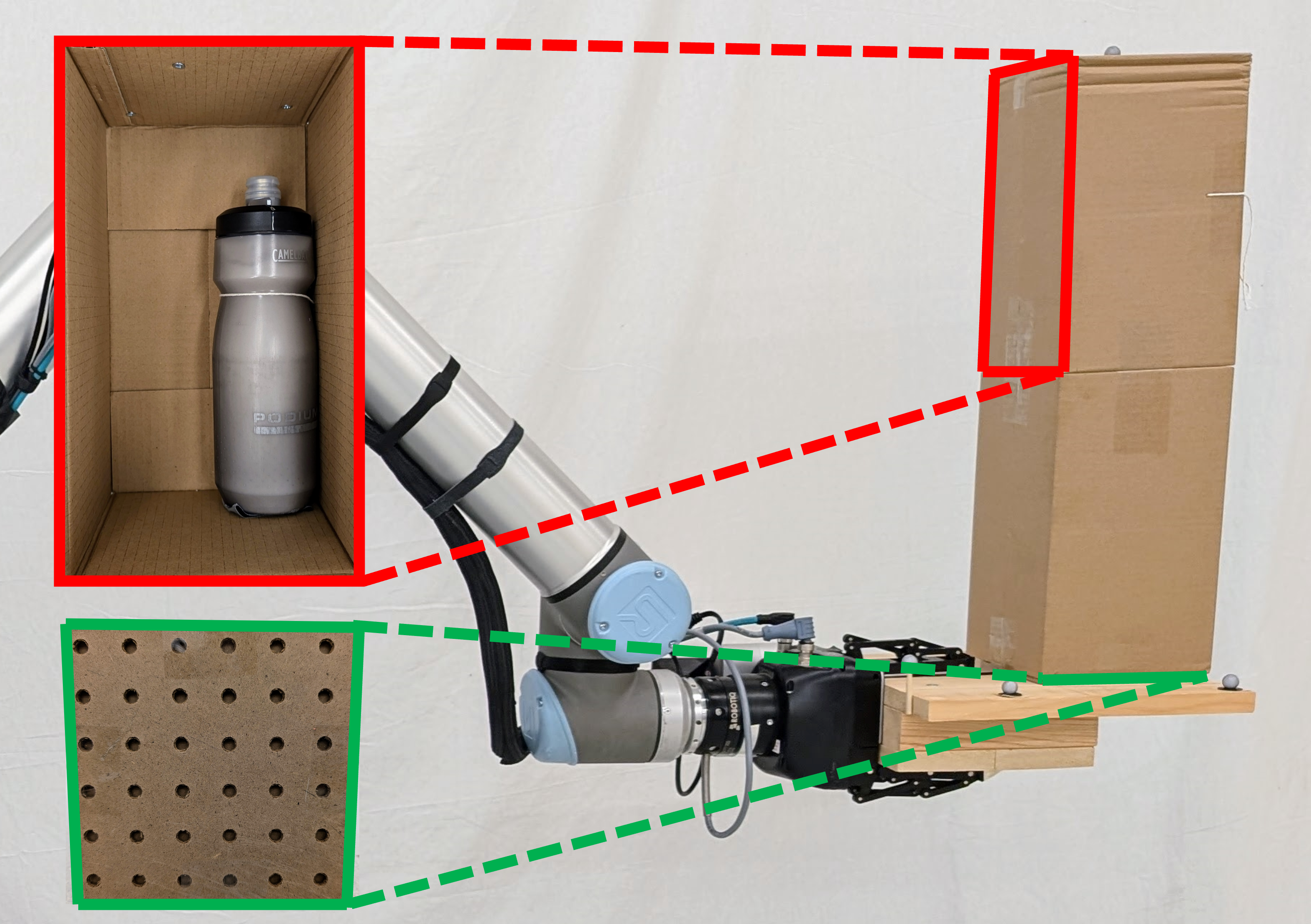}}
  \caption{Our transported objects are boxes containing a bottle filled with sugar to
  offset the CoM and to make the task of balancing more
  difficult. One cannot tell how the box is packed (and therefore what its
  mass distribution is) just by looking at it. Box2 (shown on the right)
  consists of two boxes attached together; Box1 is a single box. A firm base
  board (green) is attached to the bottom box to provide a consistent contact area
  with the tray.}
  \label{fig:box_with_bottle}
\end{figure}

We also perform experiments on our real mobile manipulator balancing a
cardboard box, as shown in Fig.~\ref{fig:eyecandy}. We test two heights of box,
Box1 and Box2, each containing a bottle filled with sugar in one corner to
offset the CoM (see Fig.~\ref{fig:box_with_bottle}). Box1 has a height
of~$h_1=\SI{28}{cm}$ and a square base with side length~\SI{15}{cm}. Its
total mass is~\SI{933}{g}, with the bottle contributing~\SI{722}{g}. Box2 is
made of two stacked boxes attached together, with the top one containing the
bottle. Its total mass is~\SI{1046}{g} and its height is~$h_2=\SI{56}{cm}$; its
base dimensions are the same as Box1. A rigid board is attached to the base of
the boxes to ensure consistent contact with the tray (again, refer to
Fig.~\ref{fig:box_with_bottle}). The friction coefficient between the boxes
(with the attached base board) and the tray was experimentally measured to
be~$\mu=0.29$. Position feedback is provided for the arm by joint encoders
at~\SI{125}{Hz} and for the base by a Vicon motion capture system
at~\SI{100}{Hz}. The laptop specifications and planner parameters are the same
as in simulation. The control loop is run at the arm's control frequency
of~\SI{125}{Hz}.

\begin{figure}[t]
  \centering
    \includegraphics[width=\columnwidth]{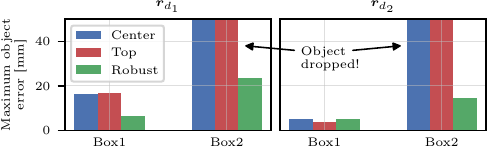}
    \caption{Maximum object displacement error for the different constraint
      methods, desired positions, and box objects. The object displacement
      error is the maximum distance the object moves from its initial position
      relative to the tray. The errors are similar between each method with
      Box1 (the shorter box), but the Center and Top baselines fail with Box2
      (the taller box), while the proposed Robust constraints successfully
      transport it.}
  \label{fig:object_error}
\end{figure}

\begin{figure}[t]
  \centering
    \includegraphics[width=\columnwidth]{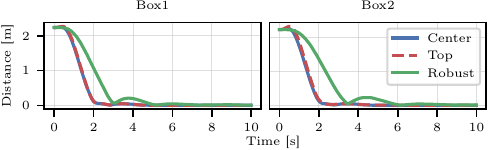}
    \caption{Distance between the EE position and the desired position over time for
      desired position~$\bm{r}_{d_1}$ and each sticking constraint method. The
      trajectories with the Center and Top constraints, in which only a single
      CoM value is considered, are nearly the same. The Robust constraints result
    in slower convergence, but always successfully transport the object.}
  \label{fig:dists_vs_time}
\end{figure}

We consider the scenario when the planner does not know how the box is packed,
and therefore its inertial parameters are not known exactly. We again test the
Center, Top, and Robust constraints using the desired positions~$\bm{r}_{d_1}$
and~$\bm{r}_{d_2}$.\footnote{We did not use the desired position~$\bm{r}_{d_3}$
because in that case if the box falls, it falls onto the robot, possibly
causing damage.} The Robust constraints assume the CoM lies in the
cuboid~$\mathcal{C}$ with dimensions~$\SI{12}{cm}\times\SI{12}{cm}\times h_i$,
where~$i=1,2$ corresponds to Box1 or Box2, which is large enough to
contain any possible centroid of the bottle no matter its position in the box.
We perform up to three runs of each combination of desired position and
constraint method; if a given combination fails before completing three
runs, we stop to avoid extra damage to the boxes. Using~$n_s=3$, each of the
constraint methods successfully transported Box1 for three runs, but only the
Robust constraints were able to do so with Box2 (with either~$n_s=3$ and~$n_s=10$).
With both values of~$n_s$, the Center constraints failed immediately
with~$\bm{r}_{d_1}$, and completed one run of~$\bm{r}_{d_2}$ before dropping
the box on the second run; the Top constraints failed immediately for both
desired positions. The maximum object displacement errors (for~$n_s=3$) are shown in
Fig.~\ref{fig:object_error}. The Robust constraints produce smaller errors with
both Box1 and Box2 (there is still \emph{some} error, due to
unmodelled effects like vibrations, drag, and inevitable model inaccuracies); the
Center and Top baselines obviously produce large errors with Box2 since the box
was dropped. The planning times were similar to those in simulation.

\begin{table}[t]
  \caption{The maximum planned EE
  velocity and acceleration as well as the root-mean-square tracking error of the
  arm's joint angles and the base's position and yaw angle in the hardware experiments.}
  \centering
  \setlength\tabcolsep{4pt}
  \begin{tabular}{l c c c c c c}
    \toprule
    & \multicolumn{3}{c}{Box1} & \multicolumn{3}{c}{Box2} \\
    \cmidrule(lr){2-4}\cmidrule(lr){5-7}
    & Center & Top & Robust & Center & Top & Robust \\
    \midrule
    EE max. vel. [\si{m/s}]   & 2.32 & 2.36 & 1.27 & 2.36 & 2.43 & 1.08 \\
    EE max. acc. [\si{m/s^2}] & 4.24 & 4.46 & 1.17 & 4.46 & 4.88 & 0.87 \\
    \midrule
    Arm err. [\si{deg}]    & 0.38 & 0.37 & 0.22 & 0.39 & 0.33 & 0.21 \\
    Base pos. err. [\si{cm}]   & 2.67 & 2.82 & 1.96 & 2.56 & 3.16 & 1.48 \\
    Base yaw err. [\si{deg}]   & 1.80 & 1.90 & 0.62 & 1.68 & 2.25 & 0.31 \\
    \bottomrule
  \end{tabular}
  \label{tab:exp_results}
\end{table}

The convergence of the EE to the desired position for~$\bm{r}_{d_1}$ is shown
in Fig.~\ref{fig:dists_vs_time} and additional metrics are in
Table~\ref{tab:exp_results}. While the Center and Top constraint methods
converge faster, this comes with the risk of dropping uncertain objects,
especially taller ones like Box2. While the Robust constraints still produce
fast motion, the maximum velocity and acceleration is reduced compared to the
baselines. The root-mean-square tracking error (RMSE) for the arm is
quite low, which suggests that neglecting the object mass was reasonable. The
RMSE for the base is higher, suggesting that adding mobility to the waiter's
problem requires additional robustness to error.

\section{Conclusion}

We present a planning framework for nonprehensile object transportation that is
robust to uncertainty in the transported object's inertial parameters. In
particular, we explicitly model and design robust constraints for uncertainty
in the object's CoM, and demonstrate successful transportation of tall objects
in simulation and on real hardware. We also use moment relaxations to develop
conditions for the inertial parameters of the transported object to be
physically realizable, which allows us to determine if the constraints would be
violated for \emph{any} possible value of the inertial parameters, including
the inertia matrix, along the planned trajectories.

\bibliographystyle{IEEEtran}
\bibliography{bibliography}

\end{document}